\documentclass[a4paper,conference]{IEEEtran}
\usepackage{times}
\usepackage{epsfig}
\usepackage{graphicx}
\usepackage{amsmath}
\usepackage{amssymb}
\usepackage{arydshln}
\usepackage{color}
\usepackage{multirow}
\usepackage{microtype}
\usepackage{subfigure}
\usepackage{booktabs}
\usepackage{amsthm}
\newtheorem{theorem}{Theorem}
\newtheorem{lemma}{Lemma}
\newtheorem{definition}{Definition}

\usepackage{algorithm}  
\usepackage{algorithmic} 
\usepackage[breaklinks=true,bookmarks=false]{hyperref}

\usepackage{xcolor}
\ifCLASSINFOpdf
\else
\fi
\begin{document}
%
\title{ODG-Q: Robust Quantization via Online Domain Generalization}

\author{Chaofan Tao,  \quad Ngai Wong\\
The University of Hong Kong \\
Email:  cftao@eee.hku.hk,  \quad nwong@eee.hku.hk
}


\maketitle

\begin{abstract}
Quantizing neural networks to low-bitwidth is important for model deployment on resource-limited edge hardware. Although a quantized network has a smaller model size and memory footprint, it is fragile to adversarial attacks. However, few methods study the robustness and training efficiency of quantized networks. To this end, we propose a new method by \textit{recasting} robust quantization as an online domain generalization problem, termed ODG-Q, which generates diverse adversarial data at a low cost during training. ODG-Q consistently outperforms existing works against various adversarial attacks. For example, on CIFAR-10 dataset, ODG-Q achieves 49.2$\%$ average improvements under five common white-box attacks and 21.7$\%$ average improvements under five common black-box attacks, with a training cost similar to that of natural training (viz. without adversaries). To our best knowledge, this work is the \emph{first work} that trains both quantized and binary neural networks on ImageNet that consistently improves robustness under different attacks. We also provide a theoretical insight of ODG-Q that accounts for the bound of model risk on attacked data.
\end{abstract}


%
\IEEEpeerreviewmaketitle


%
\IEEEpeerreviewmaketitle

\section{Introduction}
Despite the impressive performance of deep neural networks (DNNs) in various tasks \cite{chen2021litegt, tao2020dynamic, bin2019mr}, they are vulnerable to adversarial attacks~\cite{goodfellow2014explaining}. A well-trained DNN can be easily fooled to misclassify by only a small (but carefully computed) perturbation on the input data that is even not noticeable to the human eyes. Such adversarial attacks can lead to serious security and safety concerns such as autonomous driving, mobile robots, face recognition, etc. On the other hand, quantization~\cite{banner2018post, wang2019haq, tao2021fat, tao2022compression} has become an essential step for DNN deployment, especially on resource-limited edge devices. Quantization converts the data format of weights and activation from floating-point (FP) to low-bitwidth integer (INT), thereby significantly reducing the computation and storage budgets.  

Although quantization (or binarization in the extreme case) can largely reduce model deployment cost, a natural question arises: \textit{are quantized models robust enough to defend against adversarial attacks?} According to~\cite{lin2019defensive}, models are vulnerable under adversarial attacks when they are quantized to low bitwidths, e.g., $\le$ 4 bits. When quantized to 1-bit, one gets binary neural networks~\cite{rastegari2016xnor, lin2020rotated} that are extremely hardware-friendly but their robustness against attacks, especially for large-scale problems, remains underexplored. In addition, to improve model robustness, most previous defense methods~\cite{madry2017towards, dong2018boosting, zhou2020dast} based on the FP models fall short to balance efficiency and robustness. For example, training with augmented adversarial samples offline can improve robustness but typically incurs $3\times\sim30\times$ overhead vs natural training~\cite{NEURIPS2019_7503cfac}, i.e., training on clean data without defense consideration. For example, Ref.~\cite{xie2019feature} trains a robust classifier on ImageNet using 128 V100 cards at around 31$\times$ computational cost of natural training. Also, adversarial training often requires extra storage for adversarial samples, e.g., the PGD-$K$ training~\cite{madry2017towards} requires an outer loop for iterating the mini-batch data similar to natural training, plus an inner loop to generate $K$-step adversarial samples. Regularization-based method~\cite{shafahi2019label} enhances robustness by regularizing weights and/or activation, which does not require a huge training cost but their performance is often limited especially for large-scale datasets. AdvFree~\cite{NEURIPS2019_7503cfac}  uses 4 P100 GPUs and 2 days to train a FP model on ImageNet. AdvFree improves efficiency by recycling gradients that update the weights to generate adversarial perturbations. However, it still trains with each mini-batch for multiple loops per epoch to generate adversarial updates. 


\begin{figure}[t]

\centering
\includegraphics[width=3in, height=1.8in]{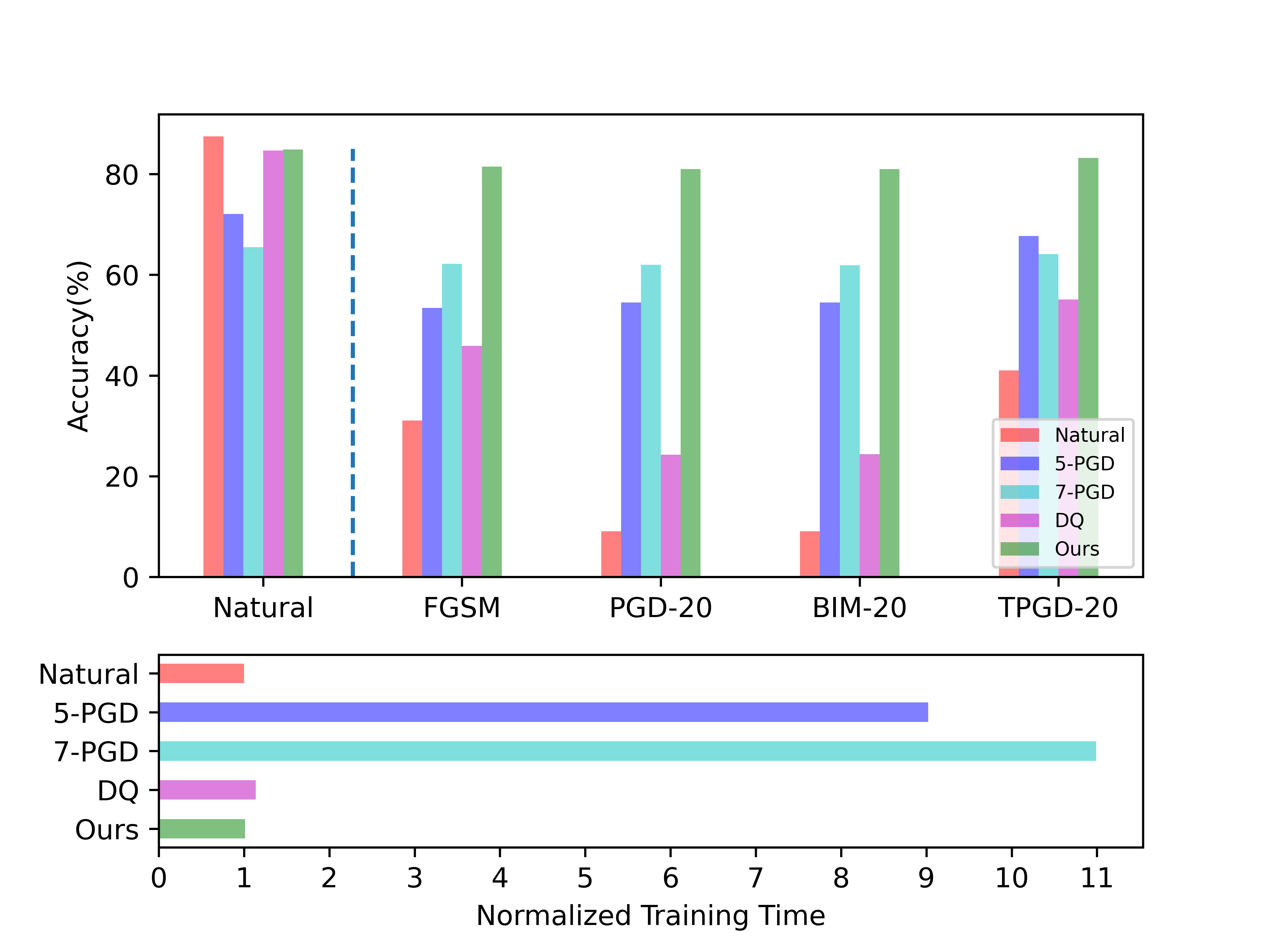}
\caption{\textbf{Right:} Comparison of defense accuracy and normalized training times on CIFAR-10 by a 4-bit ResNet-20. With a time close to natural training (41min on one NVIDIA-3090 card), our proposed method (\textcolor[RGB]{0 ,139,0}{green}) improves the performance under various attacks by a large margin compared with natural training (\textcolor{red}{red}), referred to training on
the clean data without defense.}
\label{fig:intro}
\end{figure}


We argue that the holy grail is to generate diverse adversarial samples at a very low cost. Our main innovation is to \textit{recast} the learning of adversarial robustness into a domain generalization problem. The natural samples are viewed as a subset of source data. Unlike previous defense methods~\cite{goodfellow2014explaining, madry2017towards, zhou2020dast} that only consider local adversarial perturbations generated from each sample itself in each batch, we argue that global adversarial perturbations in different batches are also important, which help the model to generate diverse adversarial samples. By constructing a set of adversarial perturbations and updating the subset of perturbations in turn during each iteration like  a roulette, we obtain a bunch of adversarial samples from multiple source domains which have different times of adversarial updating. The adversarial perturbations are generated online so that different samples arrive in a batch-sequential order without an inner loop, thereby reducing the training time sharply and obviating the need to store adversarial samples (cf. Fig.~\ref{fig:intro}). 



To sum up, the contributions of this paper are threefold. 1) We recast the learning of adversarial robustness as a domain generalization problem. By generating adversarial source data from multiple domains online during training, and enforcing high-level feature alignment, the proposed algorithm achieves state-of-the-art robustness with a runtime close to that of natural training. 2) Inspired by the theory of domain generalization, we also analytically prove that the risk on target data (attacked samples on the inference model) can be bounded with the proposed approach. 3) Extensive experiments on MNIST, CIFAR-10 and ImageNet datasets demonstrate the effectiveness of the proposed ODG-Q, under various adversarial attacks in both white-box and black-box settings.

\section{Related Work and Preliminaries}

\subsection{Objective of Robustness Learning}
\begin{definition} Given a natural dataset $(x,y)$ ($x$: data, $y$: label) sampled from a domain $D$ and a task loss function $\mathcal{L}_\Theta(\cdot,\cdot)$, the goal of learning a robust model is to learn parameters $\Theta$ that maximize the performance under any perturbation $p$ (within an $\epsilon$-radius $\ell_{\infty}$-ball) on $x$
\begin{equation}
\label{eq:def1}  
\min_\Theta \mathbb{E}_{(x,y)\sim D}[\max_p \mathcal{L}_\Theta(h(x+p),y)]~\mbox{where}~\left \|p  \right \|_{\infty} < \epsilon.%
\end{equation}%
\end{definition}%
\noindent From Eqn.~\ref{eq:def1}, we observe that optimizing the robustness of neural networks is a saddle point problem. It can be decomposed as an inner maximization problem to strengthen the deception ability of the attack given the perturbation range $\epsilon$, and an outer minimization problem to minimize the expected error given any attack.

\subsection{Representative Attacks}
FGSM~\cite{goodfellow2014explaining} is a fast attack method that performs a one-step update using the sign of the gradient of the loss function, which increases loss in its sharpest direction.
\begin{equation}
    \hat{x} = x + \epsilon \cdot \rm{sign}[\nabla_{x} \mathcal{L}_\Theta(h(x),y)].
\end{equation}
BIM~\cite{kurakin2016adversarial} performs FGSM attack in multiple steps with a small step size $\alpha$ and clips each perturbation update into a given range $\epsilon$.
\begin{equation}
    \hat{x}_{t+1} = \rm{Clip} \left \{ \hat{x}_{t} + \alpha \cdot  \rm{sign}
    \left [ \nabla_{x} \mathcal{L}_\Theta(h(\hat{x}_{t}),y) \right ] \right \}.
\end{equation}
PGD~\cite{madry2017towards} is a variant of BIM that adds a random start. It projects the updated adversarial samples into an $\epsilon$-radius $\ell_{\infty}$-ball. PGD is proven to be a general first-order attack~\cite{madry2017towards}. TPGD~\cite{zhang2019theoretically} is an improved version of PGD that considers robustness loss as the sum of the natural loss (classification loss) and the boundary loss through
\begin{subequations}
\begin{equation}
\bar{x}_t =   \rm{sign}
    \left [ \nabla_{x} \mathcal{L}_\Theta(h(\hat{x}_{t}),y)+
      \mathcal{L}_\Theta(h(\hat{x}_{t}),h(x)) / \rho  \right ],
\end{equation}
\begin{equation}
\hat{x}_{t+1} =  \rm{Clip} \left \{ \hat{x}_{t} + \alpha \cdot \bar{x}_t  \right \},
\end{equation}%
\end{subequations}%
where $\rho$ models the tradeoff between the two loss terms. These four attack schemes are employed in our experiments.

On the defense side, adversarial training~\cite{madry2017towards, tramer2017ensemble} generates adversarial examples and then trains the model with them to improve robustness. Another branch of defense is randomization, which aims to randomize the adversarial perturbation into a random perturbation. Ref.~\cite{xie2017mitigating} proposes a padding- and resize-based random input transformation to reduce adversarial perturbation. Random mask~\cite{luo2020random} filters the feature maps selectively with random masks, which helps a CNN capture spatial information and develop resistance to attacks.

\begin{figure*}[ht]
\centering
\subfigure[Representative defensive approaches]{
\begin{minipage}[t]{0.4\linewidth}
\centering
\includegraphics[width=2.in, height=1.5in]{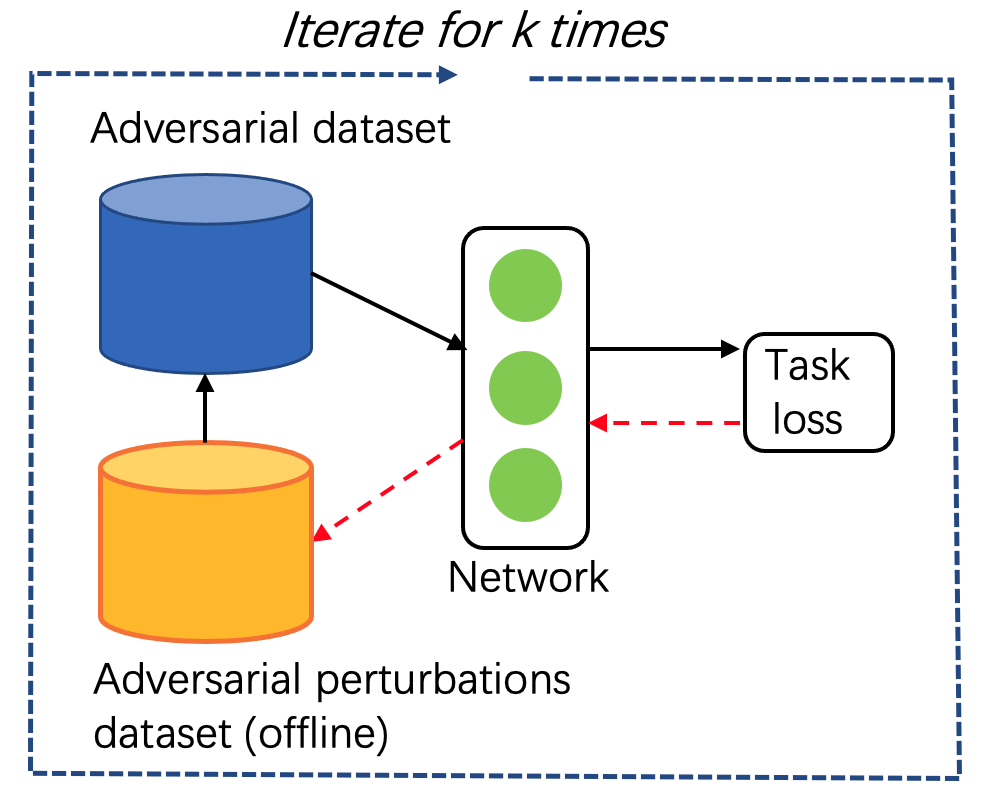}
\end{minipage}%
}%
\subfigure[Our defensive framework]{
\begin{minipage}[t]{0.6\linewidth}
\centering
\includegraphics[width=2.5in, height=1.5in]{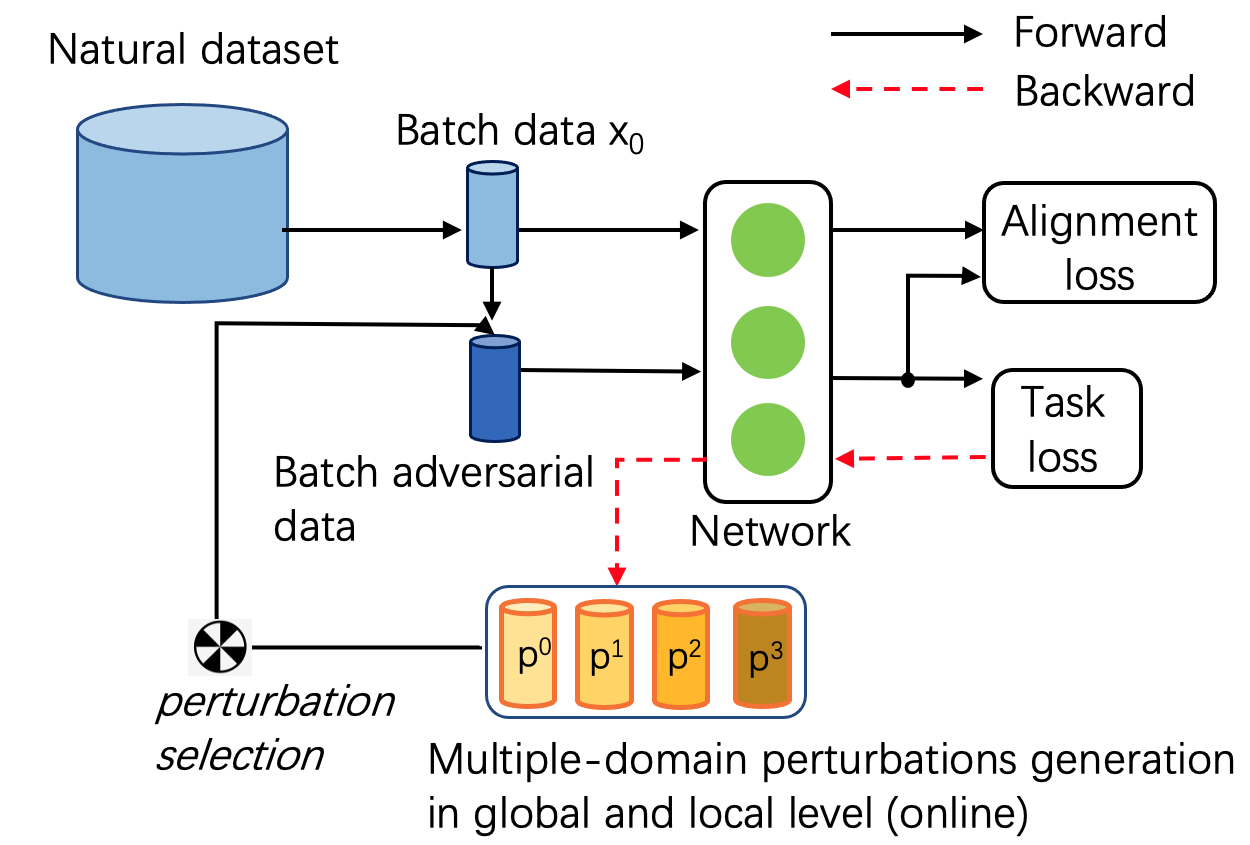}
\end{minipage}
}%
\caption{\textbf{Left:} Conventional defensive approaches are time-consuming and spatially inefficient. The model is offline that requires far more iterations than natural training to generate adversarial samples and visibility of the whole dataset. \textbf{Right:} The proposed framework instantiates a global set that updates the inter-batch perturbation and utilizes the gradient information in one backward pass for weight update and local intra-batch perturbation generation in an online manner. The alignment loss (viz. MMD) is imposed on top to minimize the discrepancy of feature distributions between natural and adversarial data.}
\label{fig:method}
\end{figure*}

\subsection{Maximum Mean Discrepancy (MMD)}
To bound the error in the target domain with assistance from source domains,  MMD~\cite{sejdinovic2013equivalence} has been extensively utilized as such a discrepancy metric.
\begin{definition}
Let $\mathcal{H}$ be the Reproducing Kernel Hibert Space (RKHS) endowed with mapping function $\phi(\cdot): X \rightarrow \mathcal{H}$. The MMD distance $d_{MMD}(\cdot,\cdot)$ represents the  distances between mean embeddings of feature generated by function $\phi(\cdot)$. The expected MMD distance between two domains $S$ and $T$ is formalized by%
\begin{equation}
\label{mmd}
d_{MMD}(S,T)=||\mathbb{E}_{(x, y) \sim S} \phi(x) - \mathbb{E}_{(x, y) \sim T} \phi(x)||_{\mathcal{H}},%
\end{equation}%
\end{definition}%
\noindent which usually utilizes induced kernel function $k(x_1, x_2)=\left \langle \phi(x_1), \phi(x_2) \right \rangle_{\mathcal{H}}$ for MMD distance computation. In this work, the adopted kernel for $\phi(\cdot)$ is the commonly used Gaussian kernel $k(x_1, x_2)=\exp(-\frac{1}{2\sigma }\left \| x_1 - x_2 \right \|^2)$ which is Lipschitz-continuous. MMD is employed to align the feature distribution between $S$ and $T$ in domain adaptation, which can be readily extended to  domain generalization.




\section{Methodology}

\begin{algorithm}[tb]
\caption{ODG-Q Training.}
\label{alg1}
\textbf{Input}: Natural dataset $(X_0, Y)$  from domain $S_0$,\\
\textbf{Parameter}: Number of online adversarial domains $N_k$, local magnitude $\epsilon_l$, perturbation bound $\epsilon$, learning rate $\eta$, trade-off factor $\lambda $, quantized model $h(\cdot)$\\
\textbf{Output}: A robust quantized model with parameter $\Theta$.
\begin{algorithmic}[1] 
\STATE Initialize global perturbation set $\mathcal{P}$ $\in R^{N_k\cdot B\cdot C\cdot H \cdot W}$ with zeros, $k=0$.
\WHILE{epoch $\leq$ $N_e/2$}
\WHILE{batch $\leq$ $N_b$}
        \STATE $p_g^k \gets \mathcal{P}[k, ...]$  
        \STATE $x_{g_k} \gets x_0+ {\rm Clip}(p_g^k, -\epsilon, +\epsilon)$
        \STATE $g_x^k \gets  \nabla_x $ $ \mathcal{L}_{task}(h(x_{g_k}), y)$
        \STATE $p_l^k \gets \epsilon_l \cdot {\rm sign}(g_x^k)$ 
        \STATE $\mathcal{P}[k:N_k-1, ...] \gets \mathcal{P}[k:N_k-1, ...] + p_l^k$  
        \STATE $x_{adv_k} \gets x_{g_k} + {\rm Clip}(p_l^k, -\epsilon, +\epsilon)$
        \STATE $g_{\Theta} \gets  \nabla_{\Theta} $ $[ \mathcal{L}_{task}(h(x_{adv_k}), y) + \lambda  \mathcal{L}_{MMD}(h(x_{adv_k}), h(x_0))]$  
        \STATE $\Theta \gets \Theta - \eta \cdot g_{\Theta}$

\ENDWHILE
\STATE $k = (k+1){\rm mod} N_k$
\ENDWHILE
\STATE \textbf{return} Robust quantized model with parameters $\Theta$.
\end{algorithmic}
\end{algorithm}


Since the adversarially augmented data introduced during the robustness training have a different distribution from the natural data, and attacked data (from target domain) is  unknown during training, it makes sense to \textit{view robustness training as domain generalization}.  To this end, we \textit{recast} the goal from a domain generalization viewpoint. Here we denote each clean datum as $(x_0, y)$ drawn from a natural source domain $S_0$ , and the augmented adversarial samples during training are drawn from other source domains $S_1, ..., S_{N_k}$. 
\begin{definition}
\label{def1}
Given $(N_k+1)$ source domains, where natural data $(x_0, y) \sim S_0$, and multiple adversarial data $(x_{adv_k}, y) \sim S_k$ for $k=1,..., N_k$. The joint distribution $P_{XY}^i \neq P_{XY}^j$ for any $1 \leq i \neq j \leq N_k+1$. The goal is to learn  robust and quantized parameters $\Theta$ that achieve minimum performance error on unseen attacked target data  $(x_t, y) \sim T$, \textit{i.e. } generated during inference,
\begin{equation}
\label{eq:def2}  
\min_\Theta \mathbb{E}_{(x_t,y)\sim T}[\max_p \mathcal{L}_\Theta(h(x_t),y)], 
\end{equation}
where the target input data $x_t$ is generated by $x_t= \mathop{\mathrm{argmax}}_p \mathcal{L}_\Theta(h(x_0+p),y),  \left \|p  \right \|_{\infty} < \epsilon $. Both model parameters $\Theta$ and activations are limited to a small bitwidth (e.g., $\leq$ 4-bit).
\end{definition}

In the following, the loss function $\mathcal{L}_\Theta(\cdot, \cdot)$ is denoted as $\mathcal{L}$ for brevity. To improve the robustness of  model against different attacks, a direct idea is to prepare  diverse adversarial samples which have different times of adversarial attack iterations, different approaches of attacks, etc. Then adversarial samples can be combined with natural samples for training. 

However, the aforementioned direct idea requires a huge space to store these adversarial samples, and expensive computation is needed for generating and training these adversarial samples. In addition, quantization/binarization introduces extra computation for all regular convolutional and fully-connected layers. To solve these problems, we propose that the adversarial data from multiple source domains $S_1, ..., S_{N_k}$ are generated and updated sequentially on-the-fly during training, instead of preparing them before training. 

As shown in Fig.~\ref{fig:method}, representative defensive approaches suffer from low time-space efficiency due to the multiple iterations offline to generate adversarial training samples. However, the visibility of the whole adversarial training dataset may not be necessary to the model. Inspired by online learning~\cite{hoi2018online}, the main drawback of offline approaches can be overcome if the adversarial training data stream can be generated online continuously.

Intuitively, the adversarial samples should meet two requirements: 1) they should be generated at a low computational cost, therefore extra training loops should be avoided; 2) they should be diverse within the $\epsilon$-radius $\ell_{\infty}$-ball of clean data. Therefore,  the gradient ascent information is considered in both global-level (inter-batch) perturbation and local-level (intra-batch) perturbation, but not  local-level perturbation only like in previous schemes~\cite{goodfellow2014explaining, madry2017towards, zhou2020dast}.

To this end, we use a global perturbation set $\mathcal{P}$ $\in R^{N_k\cdot B\cdot C\cdot H\cdot W}$, where $N_k$ is the number of online adversarial domains, and the remaining dimensions have the same shape as a batch of input data. The global perturbation set $\mathcal{P}$ accumulates the inter-batch gradient information. To make the adversarial samples diverse for producing different adversarial updates, we only update $\mathcal{P}[k:N_k-1, ...]$ per epoch, where $k$ is the epoch index modulo $N_k$. By doing so, $\mathcal{P}[0, ...]$ updates every $N_k$ epochs, and $\mathcal{P}[N_k-1, ...]$ updates every epoch like a roulette. The adversarial source datasets $\left \{ (X_{adv_k}, Y) \sim S_k | k=1,...,N_k \right \}$ are thereby constructed by incorporating global perturbation $\mathcal{P}[k:N_k-1, ...]$ and local perturbation on the natural dataset $(X_0, Y)$ sequentially.  $\mathcal{P}$ helps to recycle the gradient information from perturbations generation to weight updating.



\setlength{\tabcolsep}{0.4mm}
\begin{table*}
\footnotesize

\caption{Top-1(\%) accuracy of 4-bit ResNet-20 on CIFAR-10 against various attacks ($\epsilon=8$). Both weights and activation are quantized to low-bitwidth. ``Imp.'' and ``B'' denote average performance improvement on each attack method vs natural training and black-box attack, respectively. The last two columns respectively show the total training time (min) and per-epoch training time (s).}
\label{tab1}
\centering
\begin{tabular}{l|c|c|c|c|c|c|c|c|c|c|c|c|c|c|c}
\toprule
Method  &Natural &GN & FGSM   & PGD-20   & BIM-20& TPGD-20 &Imp.$\uparrow$ & GN(B)&FGSM(B)  & PGD-20(B) &BIM-20(B)&TPGD-20(B)&Imp.(B)$\uparrow$ & Total$\downarrow$& Epoch$\downarrow$ \\

\hline
Natural & 87.5& 72.9&31.1&9.1&9.1&41.0   & & 56.1 & 43.9 &22.8 &22.8& 50.4 & &40.5&8.1\\

\hline
5-PGD   &72.1&69.4&53.4&54.5&54.5&67.7& +27.3 & 55.0& 45.7& 44.0& 44.1& 53.3 &+9.2 &365.4 & 73.1 \\

5-BIM   &72.1&69.4&53.4&53.4&54.5&67.8&+27.0 &55.0 &45.7 &44.0 &44.1 &53.3& +9.2&364.6 & 72.9 \\

7-PGD  &65.5&63.6&62.2&62.0&61.9&64.1& +30.1& 71.2 &51.9 &43.7& 43.7& 62.8& +15.5&445.1 & 89.0 \\

AdvFree ($m=4$) &83.9&69.1&40.5&22.9&22.9&43.8&+7.2 &68.5 &41.1 &28.8 &28.8& 59.2&+6.1 &63.8 & 51.0 \\

AdvFree ($m=6$) &81.8&70.1&42.1&20.3&38.8&45.8&+10.8 &70.3 &47.7 &27.8 &25.5 &58.2&+6.7 &54.6 &65.5 \\

DQ   &84.7&73.3&45.9&24.3&24.4&55.1&+12.0 &72.1 &41.0 &29.7 &29.9 &59.9&  +7.3  &46.0 &9.2\\

\hline
Ours ($N_k=1$) &84.9&82.0&81.4& 81.0& 81.0&83.0& +49.0&82.4 &61.6 &49.4 &49.3 &67.1& \textbf{+22.8}&41.2 & 16.5\\

Ours ($N_k=2$)  &84.9&82.1&81.5&81.0&81.0& 83.1& +49.1&81.8 &61.4 & 47.5 &47.5 &66.4&+21.7 &\textbf{40.5} & \textbf{16.2}\\

Ours ($N_k=4$) &84.9&82.2&81.5&81.1&81.1&83.2& +49.1&82.4 &61.6  &46.7 &46.8 &67.1& +21.7&41.0 &16.4\\

Ours ($N_k=6$) &84.9 &82.3&81.6&81.1&81.1&83.2& \textbf{+49.2} &81.8 &61.4 &47.5 &47.4 &66.4&  +21.7 &41.3&16.5\\

\bottomrule
\end{tabular}
\end{table*}

\begin{table*}
\footnotesize

\caption{Top-1(\%) accuracy of 4-bit ResNet-20 on MNIST against various attacks ($\epsilon=8$). Both weights and activation are quantized to low-bitwidth.}
\label{tab2}
\centering
\begin{tabular}{l|c|c|c|c|c|c|c|c|c|c|c|c|c|c|c}
\toprule

Method  &Natural &GN & FGSM   & PGD-20   & BIM-20& TPGD-20 &Imp.$\uparrow$ & GN(B)&FGSM(B)  & PGD-20(B) &BIM-20(B)&TPGD-20(B)&Imp.(B)$\uparrow$ & Total$\downarrow$& Epoch$\downarrow$ \\
\hline
Natural  &99.5&99.3&82.4&63.9&63.9&79.3&  &  99.4 &97.0& 95.9& 96.0& 97.2 & &35.5 & 7.1\\

\hline
5-PGD  &99.4&99.2&99.2&98.9&98.8&99.1& +21.3&    99.1 &99.1 &99.1 &99.1 &99.1&  +2.0     &276.5 & 55.3\\

5-BIM  &99.4&99.2&99.2&98.8&98.9&99.0&+21.3  & 99.1 &99.1 &99.1 &99.1 &99.1  &  +2.0  & 272.5&54.4\\

 7-PGD  &99.3&99.3&99.2&99.1&99.1&99.2& +21.4&   99.3 &99.2 &99.2 &99.2& 99.2&     +2.1     & 372.0 &74.4\\

AdvFree (m=4)  &99.5&99.0& 89.4&64.9&64.9&73.6&+0.6&  99.4& 98.0& 97.8& 98.3&  98.0&   +1.3     &50.3& 40.3 \\

AdvFree (m=6) & 99.4&99.3&88.5&65.5&65.5&77.6& +1.5&    99.3& 98.5& 98.6& 98.7 &98.9& +1.7     &49.6&59.5\\

DQ  &99.4&99.3&84.2&65.3&65.3&79.2& +0.9&    99.4 &98.9 &99.0& 99.0& 99.2&      +2.0       &   39.0 & 7.8\\

\hline
Ours ($N_k=1$)   &99.4   &99.3&99.3&99.3&99.3&99.2&+21.5&  99.4 & 99.2& 99.3& 99.3& 99.4&  +2.2  &39.4 & 15.8\\

Ours ($N_k=2$) &99.4&99.3&99.2&99.2&99.2&99.3&+21.5&  99.5& 99.4 &99.3& 99.3 &99.5&  \textbf{+2.3}   &\textbf{39.0 }& \textbf{15.6}\\

Ours ($N_k=4$) &99.4&99.3&99.2&99.2&99.3& 99.3&+21.5&  99.5&  99.2& 99.2& 99.3& 99.4& +2.2  &39.2 & 15.7\\

Ours  ($N_k=6$)  &99.4&99.4&99.3&99.3&99.3&99.3&\textbf{+21.6}&  99.4 &99.0& 99.2& 99.1& 99.2& +2.1  &39.4 & 15.8\\

\bottomrule
\end{tabular}
\end{table*}

The proposed training process is summarized in Algorithm~\ref{alg1} dubbed ``ODG-Q'' wherein we efficiently generate diverse adversarial samples online. For each batch, in Stage 1 (lines 4$\sim$7), the global perturbation (inter-batch) in domain $k$ is added to natural data to obtain $x_{g_k}$, then the local perturbation (intra-batch) is generated by gradient ascent on $x_{g_k}$. In Stage 2 (lines 8$\sim$9), the global perturbation set is updated with the current local perturbation, and the adversarial data $x_{adv_k}$ is generated using global perturbation $p_g^k$ and local perturbation $p_l^k$. In Stage 3 (lines 10$\sim$11), the model is trained with $x_{adv_k}$. An MMD loss is added on the last convolution output for high-level feature alignment. Weight updates only once   each batch without inner loop for adversarial samples generation. We halve the training epochs to keep our total training time consistent to natural training, since two backward passes are involved per batch in ODG-Q for perturbation generation and gradient descent, respectively.

The theoretical insight of ODG-Q, which discusses \textit{how the risk on attacked data during inference (data from target domain) can be bounded by the proposed algorithm}, is described in Appendix.

\section{Experiments}

\setlength{\tabcolsep}{1.7mm}
\begin{table*}
\caption{Performance of ResNet-18 on ImageNet against attacks ($\epsilon=1$ / $\epsilon=2$). Both weights and activation are quantized to low-bitwidth. The last two columns report the total training time (h) and per-epoch training time (min), respectively. }
\label{tab3}
\centering
\begin{tabular}{l|c|c|c|c|c|c|c|c|c|c|c}
\toprule
Defense Method  & Natural &  PGD-20 & BIM-20 & TPGD-20&Imp.$\uparrow$&PGD-20(B) & BIM-20(B) & TPGD-20(B)&Imp.(B)$\uparrow$ &Total$\downarrow$&Epoch$\downarrow$ \\ 
\hline
Natural 4-bit & 69.9&  40.5/17.5  &   40.4/17.5 & 55.7/38.6 &&  61.8/55.3  &61.8/55.2 & 65.7/63.4 & & 38.4&25.6\\
\hline
Ours ($N_k=1$)&  69.9&43.6/21.2   & 43.6/21.4&  59.3/42.1& +3.4/+3.7 & 63.4/58.6 &63.4/58.6 &68.4/66.0 & +2.2/+3.1& 42.7 & 56.9 \\
Ours ($N_k=2$) &69.8& 43.5/20.9&43.5/20.9&59.4/42.1& +3.3/+3.5 &   63.8/58.9 &63.8/59.0 &68.3/65.8& +2.1/+3.3&  43.1 &57.5\\
Ours ($N_k=4$)  & 69.9& 43.8/21.3& 43.7/21.3&59.3/42.4&\textbf{+3.5/+3.8} &63.2/58.1 &63.1/58.0 &68.1/65.4 &+1.7/+2.5    & \textbf{42.7} &\textbf{56.9}\\ 
Ours ($N_k=6$)& 69.9& 43.6/21.4& 43.7/21.3& 59.3/42.4&+3.4/+3.9 & 64.4/60.0 &64.3/60.0 & 68.6/66.3& \textbf{+2.7/+4.1} & 42.8 & 57.1\\
\hline
Natural 1-bit &  48.9 &  41.3/37.2 &  41.0/37.3  & 44.9/42.1& &   34.6/26.7 & 34.6/26.7&  41.6/37.7 & &36.5&24.3\\
\hline
Ours ($N_k=1$) &  47.3& 44.0/41.2 & 44.0/41.2 & 46.5/44.0& +2.4/+3.3&  36.7/28.8& 36.7/28.9& 44.3/40.2 & +2.3/+2.3 & 40.2 & 53.6\\
Ours ($N_k=2$)&  47.2& 44.5/41.9 & 44.7/41.9 & 46.7/44.9  & \textbf{+2.9/+4.0} &  36.0/28.5 & 36.0/28.5  &43.5/39.2 & +1.6/+1.7& 40.3& 53.8\\
Ours ($N_k=4$) & 47.3 &   44.2/41.9 &  44.3/41.8 &  46.0/44.5& +2.4/+3.9&36.1/27.7  &36.1/27.7 &43.9/39.6&  +1.9/+1.3 &   \textbf{39.9} & \textbf{53.3}  \\
Ours ($N_k=6$)  & 47.3 &  44.2/41.6&  44.3/41.4&  46.7/44.3 & +2.7/+3.6 &  36.9/28.9  & 36.9/28.9&  44.1/40.2 &  \textbf{+2.4/+2.3} & 40.0&53.3\\
\bottomrule
\end{tabular}
\end{table*}

We conduct experiments on MNIST~\cite{lecun1998gradient}, CIFAR-10~\cite{krizhevsky2009learning} and ImageNet~\cite{deng2009imagenet}. Quantization-aware training on uniform quantizer and XNOR-net~\cite{rastegari2016xnor} are employed for quantization and binarization (viz. 1-bit quantization), respectively. We quantize both weights and activation.  We use different attacks for robustness evaluation, namely, FGSM~\cite{goodfellow2014explaining}, PGD~\cite{kurakin2016adversarial}, BIM~\cite{madry2017towards} and TPGD~\cite{zhang2019theoretically}. ``Natural'' refers to the evaluation of the clean natural data. ``GN'' stands for Gaussian noise. PGD-20 means the number of perturbation updates is 20. The notation is similar for BIM-20 and TPGD-20, etc. ResNet-20~\cite{he2016deep} is employed in MNIST and CIFAR-10. For ImageNet, we employ ResNet-18. Implementation details are given in Appendix.

\subsection{Experimental Results}

\setlength{\tabcolsep}{0.9mm}
\begin{table*}[t]
\footnotesize

\caption{Ablation study for different variants of our model. Top-1(\%) accuracy of 4-bit ResNet-20 on CIFAR-10 against various adversarial attacks ($\epsilon=8$). The mark ``global pert.'' means ODG-Q removes the inter-batch global perturbation, and ``MMD (0.3)'' means we replace the coefficient of MMD loss $\lambda$ from 3.0 to 0.3.}
\label{tab-ablation}
\centering
\begin{tabular}{l|c|c|c|c|c|c|c|c|c|c|c|c|c}
\toprule
Method  &Natural &GN & FGSM   & PGD-20   & BIM-20& TPGD-20 &Imp.$\uparrow$ & GN(B)&FGSM(B)  & PGD-20(B) &BIM-20(B)&TPGD-20(B)&Imp.(B)$\uparrow$  \\

\hline
Ours ($N_k=4$) &84.9&82.2&81.5&81.1&81.1&83.2& &82.4 &61.6  &46.7 &46.8 &67.1&  \\
\hline
Ours w/o global pert. &77.1    &75.1 &77.1 &77.1 &77.1 &76.9 &   -5.1        &75.1 &51.1 &46.8 &46.6 &63.9 &  -4.2 \\
 
Ours w/o MMD &88.7       &76.6 &40.2  &18.8 &18.9 &49.2 &  -41.1  &76.6 &53.6 &39.8& 39.9 &62.5& -6.4  \\
 
Ours w/ MMD (0.3) &86.5       &81.7& 66.1 &59.2 &59.2 &73.7    & -25.8    &  81.7& 53.7 &36.5 &36.8 &60.6&  -7.1 \\

\bottomrule
\end{tabular}
\end{table*}

\setlength{\tabcolsep}{1.2mm}
\begin{table}[!htb]
\scriptsize
\caption{Performance comparison between natural training and ours ($N_k=4$) with different bitwidths on CIFAR-10 dataset.}
\label{tab:more_bits}
\centering
\begin{tabular}{l|c|c|c|c|c|c|c|c}
\toprule
Bit & Method  &Natural & GN&FGSM  & PGD-20  & BIM-20& TPGD-20 & Imp.$\uparrow$ \\
\hline
\multirow{2}{*}{32bit} &Natural&88.0 &76.1&42.3&21.2&21.2&49.8&\\
~&Ours &88.0&82.1& 82.1& 82.5&82.5&83.8&\textbf{+40.5}\\
\cdashline{1-9}
\multirow{2}{*}{8bit} &Natural &87.9&73.5&41.1&18.7&18.7&47.4\\
~&Ours &  84.5&  81.9&81.3& 81.5& 81.5&83.3 &\textbf{+42.0}\\
\cdashline{1-9}
\multirow{2}{*}{4bit} &Natural&87.5& 72.9&31.1&9.1&9.1&41.0\\
~&Ours  &84.9&82.2&81.5&81.1&81.1&83.2&\textbf{+49.2}\\
\cdashline{1-9}
\multirow{2}{*}{3bit} &Natural &87.0&71.3&39.9&17.8&17.8& 44.4\\
~&Ours &84.5&81.9&81.9& 81.4&81.4&83.6&\textbf{+43.8}\\
\cdashline{1-9}
\multirow{2}{*}{2bit} &Natural  &86.0& 75.3&46.8&24.4&24.4& 48.8\\
~&Ours &  82.7&79.6&78.9&78.3&78.2&81.2&\textbf{+35.3}\\
\cdashline{1-9}
\multirow{2}{*}{1bit} &Natural &81.8&64.6&27.9&19.3&19.3&26.6&\\
~&Ours  &81.2&64.8&55.1&55.3&55.3&58.9&\textbf{+26.3} \\
\bottomrule
\end{tabular}
\end{table}

\begin{figure*}[!htb]
\centering
\subfigure[$N_k=2$ Multi-domain Online Samples Visualization]{
\begin{minipage}[t]{0.5\linewidth}
\centering
\includegraphics[width=2.9in, height=1.9in]{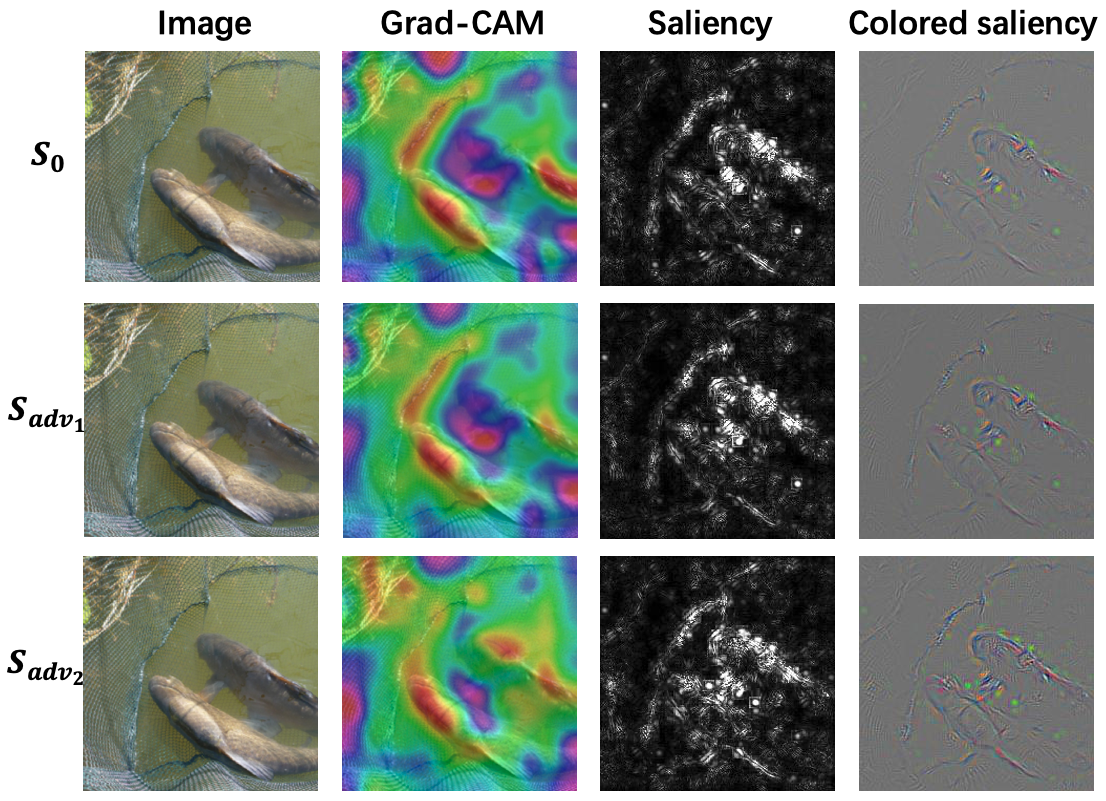}
\label{figvis_samples_a}
\end{minipage}%
}%
\subfigure[T-SNE visualization of feature distributions]{
\begin{minipage}[t]{0.5\linewidth}
\centering
\includegraphics[width=2.7in, height=1.9in]{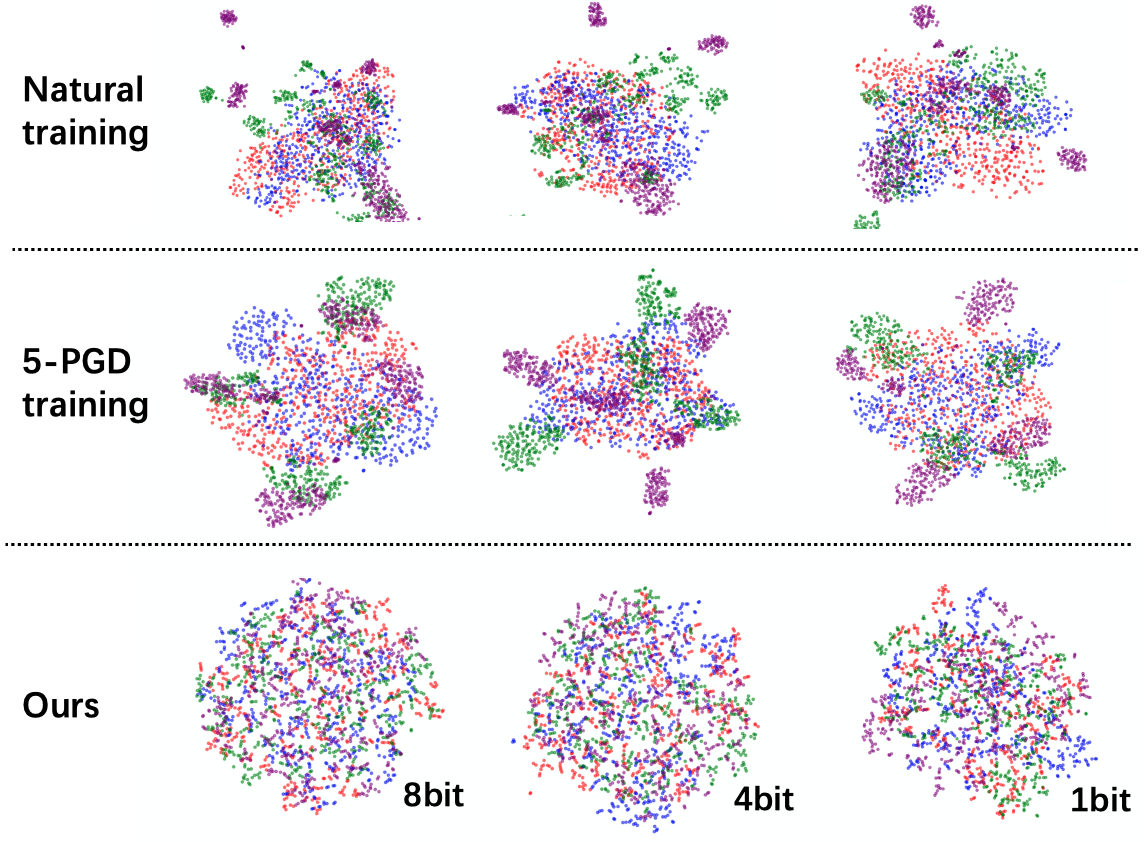}
\label{figvis_samples_b}
\end{minipage}
}%
\caption{\textbf{Left:} Visualizing what the model learns in natural domain $S_0$ and online adversarial domains  $S_{adv_0}$ and $S_{adv_1}$. The true label of example is tench (a kind of fish). 1) All the natural and adversarial data focus on the area containing  the region of interest (RoI) successfully. 2) Fewer high-confidence areas (warm color) focus on the areas without RoI in adversarial domains than that in the natural domain $S_0$. \textbf{Right:}  T-SNE visualization~\cite{van2008visualizing} of features for 
natural images (\textcolor{red}{red}), FGSM-attacked images (\textcolor{blue}{blue}), PGD-attacked images (\textcolor[RGB]{0 ,139,0}{green}) and TPGD-attacked images (\textcolor[RGB]{145,44,238}{purple}). Regardless of bitwidths, the features have large discrepancies in natural training, while being aligned well in ours, which shows why our method enables robust performance on the attacked images.
}

\label{fig:vis_samples}
\end{figure*}


Table~\ref{tab1} reports the model accuracy under both white-box and black-box attacks, together with training time for different methods on CIFAR-10.  The ``Natural'' in the first row means experiments conducted on the clean natural data, and ``Natural'' in the method column means natural training without any defense approach. From Table~\ref{tab1}, we observe that low-bitwidth neural networks are sensitive to adversarial attacks for CIFAR-10  dataset. Even Gaussian noise on the natural data causes performance degradation from 87.5\% to 72.9\%.  K-iterative Training can improve robustness. However, the expensive computational cost makes it hard to scale to large datasets. AdvFree($m=4$) and AdvFree($m=6$) are two variants of AdvFree that set 4 and 6 times inner loops, respectively. The results of AdvFree show that it may not improve robustness effectively in the low-bitwidth setting. DQ is a regularization-based method having a small computational overhead, while its performance of robustness is not as good as training with the adversarial samples.  

In contrast, our method 1) outperforms state-of-the-art methods in both accuracy and training time consistently by a large margin: achieving 49.2$\%$ average improvements under five white-box attacks and 21.7$\%$ under five black-box attacks in the setting of $N_k=6$.; 2) maintains good performance on the natural data as well. The good generalization ability on the original natural data relies on the involvement of source feature during the whole training process. Therefore, we keep sending natural data and  using MMD loss to align features between natural data and adversarial samples during training.  From the comparison from $N_k=1$ to $N_k=6$ in our approach, the robustness against attacks is improved by introducing diverse adversarial source data. $N_k$ reflects the involvement frequency of global perturbation to generate adversarial source samples, apart from local perturbation. It shows the incorporation of multiple adversarial source domains can improve robustness very effectively.

The performance on MNIST is reported in Table \ref{tab2}. Our method achieves good accuracy under various attacks, with efficient training time per epoch.
In Table~\ref{tab3}, we report the defense performance on the large-scale ImageNet dataset in both 4-bit and 1-bit settings. Due to the high computational requirement, few teams have attempted the adversarial robustness on the ImageNet dataset, let alone training in the low-bitwidth setting which also requires training extra quantization layers. \emph{To our best knowledge, this paper is the first to train both quantized and binary neural networks on ImageNet successfully against adversarial attacks.} As shown in Table~\ref{tab3}, our method boosts the robustness to PGD, BIM and TPGD attacks steadily, with the total training time close to that of natural training. Since both the adopted quantization/binarization method  and the model architecture (ResNet-18) in this paper are basic, there exists room for improvement for the ImageNet dataset.

%

\section{Discussion}
\subsection{Ablation Study of ODG-Q}
\label{dis:ablation}
We provide an ablation study of ODG-Q in Table \ref{tab-ablation}. In the setting ``w/o global pert.'', we remove the global perturbation in Algorithm \ref{alg1}. The training is still in an online version that adversarial samples are generated dynamically during training. However, only local perturbation generated from the data itself can be used for adversarial training, which is also adopted in previous methods~\cite{NEURIPS2019_7503cfac, goodfellow2014explaining, madry2017towards, zhou2020dast}. The performance gap between  ``w/o global pert.'' and the full ODG-Q demonstrates the importance of global perturbation from different batch data, which gives the model a holistic view of data and helps the model to generate diverse adversarial samples in different source domains, \textit{i.e.} the domain of source data depends on the times of using and updating global perturbation set during training. The visualization of different source data is shown in Fig.~\ref{figvis_samples_a}. The visualization  shows that the data from different source domains have different gradient information.

In the remaining settings, we remove the MMD loss or set a small MMD loss coefficient.  We can observe that MMD loss plays a significant role during training.  Without the feature alignment between natural samples and various online adversarial samples by MMD loss, the training may become unconstrained.

\subsection{Performance in Different Bitwidths}
We further test the performance of our method when quantizing the network to different bitwidths. Although this paper focuses on robustness improvement for quantized and binary models, the proposed Algorithm~\ref{alg1} is orthogonal to the used bitwidth. Therefore, we also experiment on full-precision model (32-bit), besides low-precision settings ranging from 8-bit down to 1-bit. From Table~\ref{tab:more_bits}, our approach promotes robustness against different adversarial attacks by a large margin. The robustness improves significantly not only at full-precision settings, but more importantly, in quantized and binarized settings. We also observe that the level of improvement increases from 32-bit to 4-bit, and decreases from 4-bit to 1-bit. It indicates that the robustness improvement becomes a little easier when quantized to 8-bit, demonstrating that small adversarial perturbations can be absorbed in this case. However, in a low-bitwidth setting ($\leq$ 4-bit), robustness learning becomes very challenging with a tightened bitwidth. It is reasonable since the capability of a model is strictly constrained by the low-bitwidth setting such that even training the model with diverse adversarial samples, the performance is limited. 

\subsection{Interpretability of Proposed Robust Training}
In Fig.~\ref{figvis_samples_a}, we visualize what the model has learned from the source domain $S_0$ and online adversarial domains $S_{adv_1}$ and $S_{adv_2}$. With small perturbations on the natural images, Grad-CAM~\cite{selvaraju2017grad} and Backpropagation Saliency map~\cite{simonyan2013deep} are adopted to check the attention captured by the model. Interestingly, we find that there are fewer high-confidence areas that focus on the areas without the region of interest (RoI) in adversarial domains than that in the natural domain $S_0$. For example, the Grad-CAM in $S_0$ treats not only the fishes themselves but also surrounding water areas as high-confidence areas, while the Grad-CAM in $S_{adv_1}$  and $S_{adv_2}$ tends to have lower interest in the surrounding water area. It hints that training with the online adversarial samples improves robustness by encouraging the model to focus more on the useful RoI.

We visualize the T-SNE projected features distribution of natural training, 5-PGD training and the proposed ODG-Q on CIFAR-10 dataset in Fig.~\ref{figvis_samples_b}. It is observed that the features of natural images (\textcolor{red}{red}) is quite different from the features of various attacked images, thereby the predictor using natural training can be deceived to give incorrect predictions on attacked images. Instead, by training natural domains and online adversarial domains altogether with MMD loss, the high-level features from different domains are aligned closer together, thus allowing the proposed model to give correct predictions regardless of the attacks.

\section{Conclusions}

In this paper, we novelly recast robustness training as a domain generalization problem in an online version where multiple source data are generated dynamically. We then propose an strategy to generate diverse source data considering both global and local perturbations. Powered by the MMD loss, the training process is safely constrained that high-level features of the natural and adversarial online data are aligned, which reduces the expected risk of model on attacked data during inference in theory. Extensive experiments show that the proposed ODG-Q defends various attacks with a big margin over existing defense schemes in both white-box and black-box settings, using only minimal computation and a training time near natural training. We are hopeful that this work bridges the gap between network quantization and adversarial robustness.

\section*{Acknowledgements}
This work is supported in part by the General Research Fund (GRF) project 17206020, and in part by ACCESS, AI Chip Center for Emerging Smart Systems, Hong Kong SAR.

{\small
\bibliographystyle{plain}
\bibliography{egbib}
}
\clearpage
\appendix
\section{Appendix}

\section{Theoretical Insight}
Regarding the generalization bound on the attacked target domain, a robust quantized model is supposed to constrict the upper bound of target risk. Here we discuss \textit{how the target risk can be bounded by the proposed algorithm}. 

According to the  theory of domain adaptation in \cite{ben2010theory, ben2007analysis, redko2017theoretical}, let $\mathcal{H}$ be a RKHS with kernel function $k(\cdot, \cdot)$ induced by mapping function $\phi(\cdot): X \rightarrow \mathcal{H}$. Then, the  expected risk  can be defined as 
the probability  that a hypothesis $h$ disagrees with any labeling function $f$ within the domain $D$, \textit{i.e.}, $R^D[h,f]=\mathbb{E}_{(x,y)\sim D} [\mathcal{L}(h(x), f(x))]$. We use $R^D[h]$ to abbreviate $R^D[h,f]$. 

\begin{lemma}
\label{lemma1}
For any source domain $S_k$ with $ k \in [0, N_k]$ and the unseen target domain $T$, let $\mathcal{F}= \left \{ f \in \mathcal{H}: ||f||_{\mathcal{H}} 
 \leq 1 \right \}$ be a hypothesis family inside a unit ball in RKHS $\mathcal{H}$ with a kernel $k(\cdot,\cdot)$. Assume the loss function $\mathcal{L}$ is convex with form $|h(x)-f(x)|^q$ for some $q>0$, and obeys the triangle inequality, and bounded by $||\mathcal{L}|||_{{\mathcal{H}}^q} \leq 1$. The mapping $\phi(\cdot)$ computed in the MMD distance is Lipschitz-continuous. Then, for every hypothesis $h$, we can always find a positive $\lambda_k$
\begin{equation}
\label{eq:main_lemma1}
R^{T}[h] \leq R^{S_k}[h] + \lambda_k d_{MMD}(S_k, S_0),
\end{equation}
where $S_0$ is the natural source domain and $ \left \{ S_1,...,S_{N_k} \right \}$ are $N_k$ adversarial source domains.
\end{lemma}

\begin{theorem}
\label{theo1}
Given a natural source domain $S_0$ and $N_k$ adversarial source domains $ \left \{ S_1,...,S_{N_k} \right \}$, if the assumptions in Lemma \ref{lemma1} hold, then the expected risk on the attacked target domain $T$ is bounded for some positive $\lambda$ by
\begin{equation}
\label{eq:main_theo1}
R^{T}[h] \leq \frac{1}{N_k} \sum_{k=1}^{N_k} R^{S_k}[h] + \lambda d_{MMD}(S_k, S_0),
\end{equation}
where the input data $x$ sampled from all domains $\left\{ S_0,...,S_{N_k}, T \right \}$ is around the corresponding natural data $x_0$ within a perturbation  $\epsilon$.
\end{theorem}
Eqn.~\ref{eq:main_theo1} is very similar to the optimization objective proposed in ODG-Q (line 13), in which both the task loss (cross-entropy) on the adversarial samples from multiple online source  domains and high-level feature alignment loss is minimized during training.
\\

\textbf{Proof for the Theoretical Insight}
\label{ap:proof}


\textit{Proof of Lemma \ref{lemma1}:} By assuming $||\mathcal{L}|||_{{\mathcal{H}}^q} $ is bounded by 1,  the expected risk on a domain can be measured in terms of the inner product in the corresponding RKHS \cite{redko2017theoretical}. Powered by the reproducing mapping property in RKHS, the expected risk on one domain $D$ can be re-written as 
\begin{equation}
R^D[h]=\mathbb{E}_{(x,y)\sim D} [\mathcal{L}(h(x), f(x))]=\mathbb{E}_{(x,y)\sim D}[\left \langle \phi(x), \mathcal{L}  \right \rangle_{\mathcal{H}}],
\end{equation}
which holds for domain $T$ and  $S_k, k \in [0, N_k]$. The  $||\mathcal{L}|||_{{\mathcal{H}}^q}$ in bound assumption can also be safely extended by $||\mathcal{L}|||_{{\mathcal{H}}^q} \leq n $ \cite{mansour2009domain}. Mainly followed by the domain adaptation theory \cite{redko2017theoretical} of the relationship between source and target domain, we have
\begin{equation}
\begin{aligned}
\scriptsize
& R^{T}[h] = R^{S_k}[h] + R^{T}[h] -  R^{S_k}[h] \\
          &= R^{S_k}[h] + \mathbb{E}_{(x,y)\sim T}[\left \langle \phi(x), \mathcal{L}  \right \rangle_{\mathcal{H}}] - 
          \mathbb{E}_{(x,y)\sim S_k}[\left \langle \phi(x), \mathcal{L}  \right \rangle_{\mathcal{H}}]\\
&= R^{S_k}[h] +   ||\left \langle \mathbb{E}_{(x,y)\sim T}[\phi(x)] -\mathbb{E}_{(x,y)\sim S_k}[\phi(x)],\ \mathcal{L} \right \rangle||_{\mathcal{H}}\\
&\leq R^{S_k}[h] +  ||\mathcal{L}||_{\mathcal{H}} \ || \mathbb{E}_{(x,y)\sim T}[\phi(x)] -\mathbb{E}_{(x,y)\sim S_k}[\phi(x)]||_{\mathcal{H}}\\
&\leq  R^{S_k}[h] + d_{MMD}(T, S_k) = R^{S_k}[h] + d_{MMD}(S_k, T),
\label{eq:inter}
\end{aligned}
\end{equation}
where the second line uses rewrittened expected risk, the third line uses property of expected value, and the fourth line uses Cauchy Schwartz's Inequality. Here $R^{T}[h]$ reflects the expected risk of model on attacked data during inference, and $R^{S_k}[h]$ reflects the expected risk of model on online adversarial samples in $k$-th source domain during training.

We denote the natural data as $x_0 \sim S_0$, adversarial source data from $k$-th source domain $x_{adv_k} \sim S_k$, and attacked target data as $x_t \sim T$. Based on the perturbation assumption, every $x_{adv_k}$ and $x_t$ is constricted in the $\epsilon$-radius $\ell_{\infty}$-ball around $x_0$. Hence, there exists positive $\lambda'$ that sufficiently satisfies $||x_k-x_t|| \leq \lambda' ||x_k-x_0||$. In addition, since the mapping $\phi(\cdot)$ computed in MMD distance is Lipschitz-continuous, we can always find a positive $\lambda_k$ satisfying $d_{MMD}(S_k, T) \leq \lambda_k d_{MMD}(S_k, S_0)$ that bounds $S_0$, $S_k$ and $T$ in the RKHS. Combined with Eq.\ref{eq:inter}, we have
\begin{equation}
R^{T}[h] \leq R^{S_k}[h] + \lambda_k d_{MMD}(S_k, S_0).
\end{equation}


\textit{Proof of Theorem \ref{theo1}}   Based on Lemma \ref{lemma1}, by averaging the Eq.\ref{theo1} for any $k \in \left \{1,...,N_k\right \}$, and assigning $\lambda = \max \left \{ \lambda_1, ..., \lambda_{N_k} \right \}$, we have

\begin{equation}
\label{eq:rt}
\begin{aligned}
R^{T}[h] &\leq  R^{S_k}[h] + \lambda_k d_{MMD}(S_k, S_0) \\
&\leq \frac{1}{N_k}  \left \{ \sum_{k=1}^{N_k} [R^{S_k}[h] + \lambda_k d_{MMD}(S_k, S_0)] \right \} \\
&  \leq \frac{1}{N_k} \sum_{k=1}^{N_k} R^{S_k}[h] + \lambda d_{MMD}(S_k, S_0).
\end{aligned}
\end{equation}
From the equation above, the expected risk on the target domain (attacked data generated during inference) is highly related to the minimization objective of the  proposed Algorithm in the main paper (line 10 of Algorithm 1). Since the target data are unseen during training, the MMD loss between the natural source domain and target domain cannot be directly employed. Fortunately, all the adversarial data are bounded within a certain range $\epsilon $ around the natural data. By bridging natural data and attacked data with diverse adversarial samples that have different times of adversarial updates, the expected risk on the target domain can be effectively reduced theoretically and empirically.

\begin{figure*}[t]
\centering
\includegraphics[width=5.5in, height=2.8in]{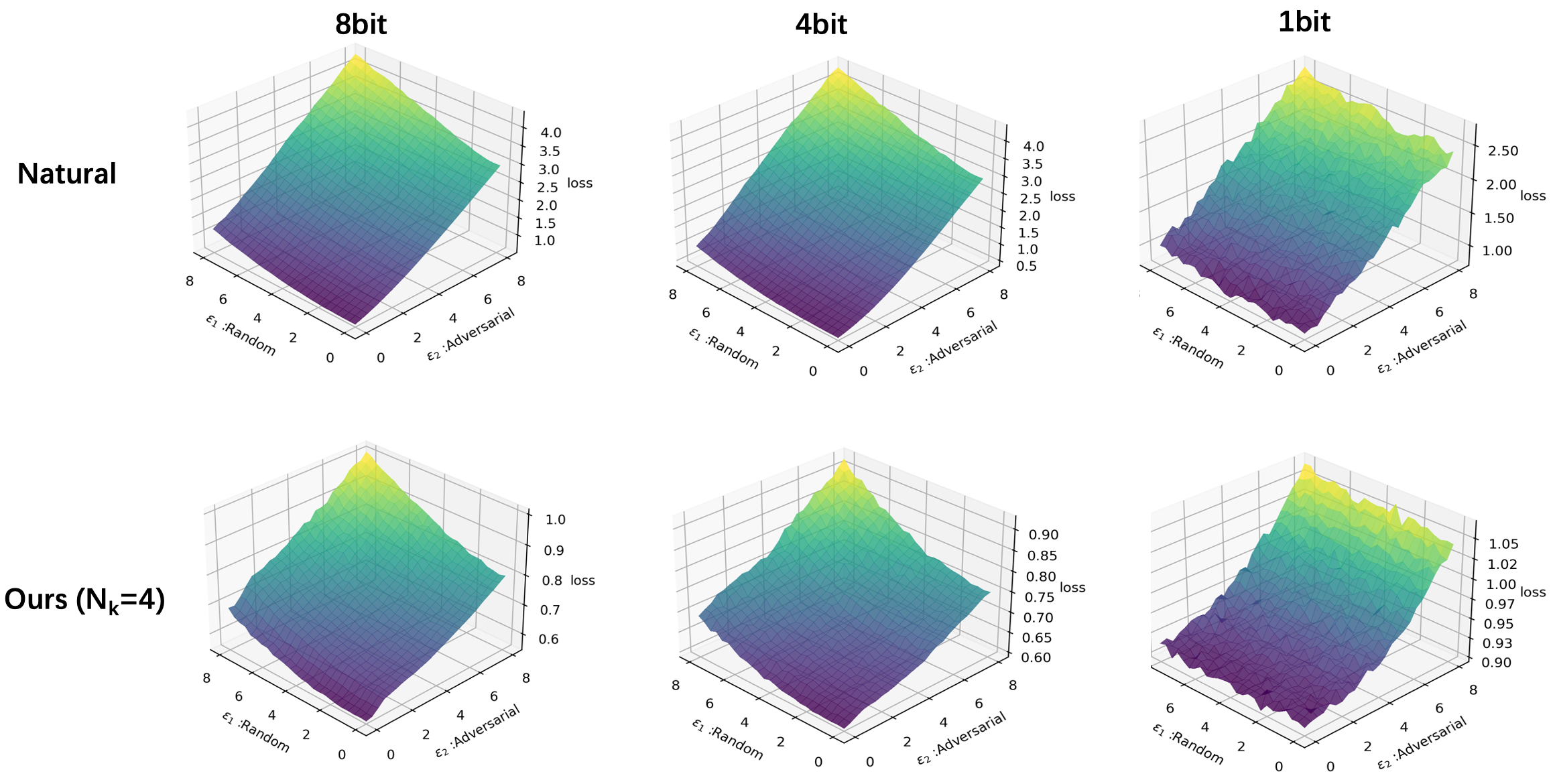}
\caption{Comparison of cross-entropy loss surfaces between natural training and ours ($N_k=4$) in 8-bit, 4-bit and 1-bit neural networks. We perturb one input image along random direction $\epsilon_1$ and adversarial direction $\epsilon_2$. We can observe that 1) the loss scale is consistently small in our approach (amplitude reduces to 1); 2) in contrast to the sharp loss that increases along the adversarial direction in natural training, the loss surface in ours increases slowly along with two directions; 3) robustness training in 1-bit case becomes even more difficult than that in the 8-bit and 4-bit cases, expectedly.} 
\label{fig:cls_loss_surface}
\end{figure*}

\begin{figure*}[th]
\centering
\includegraphics[width=5.5in,height=2.8in]{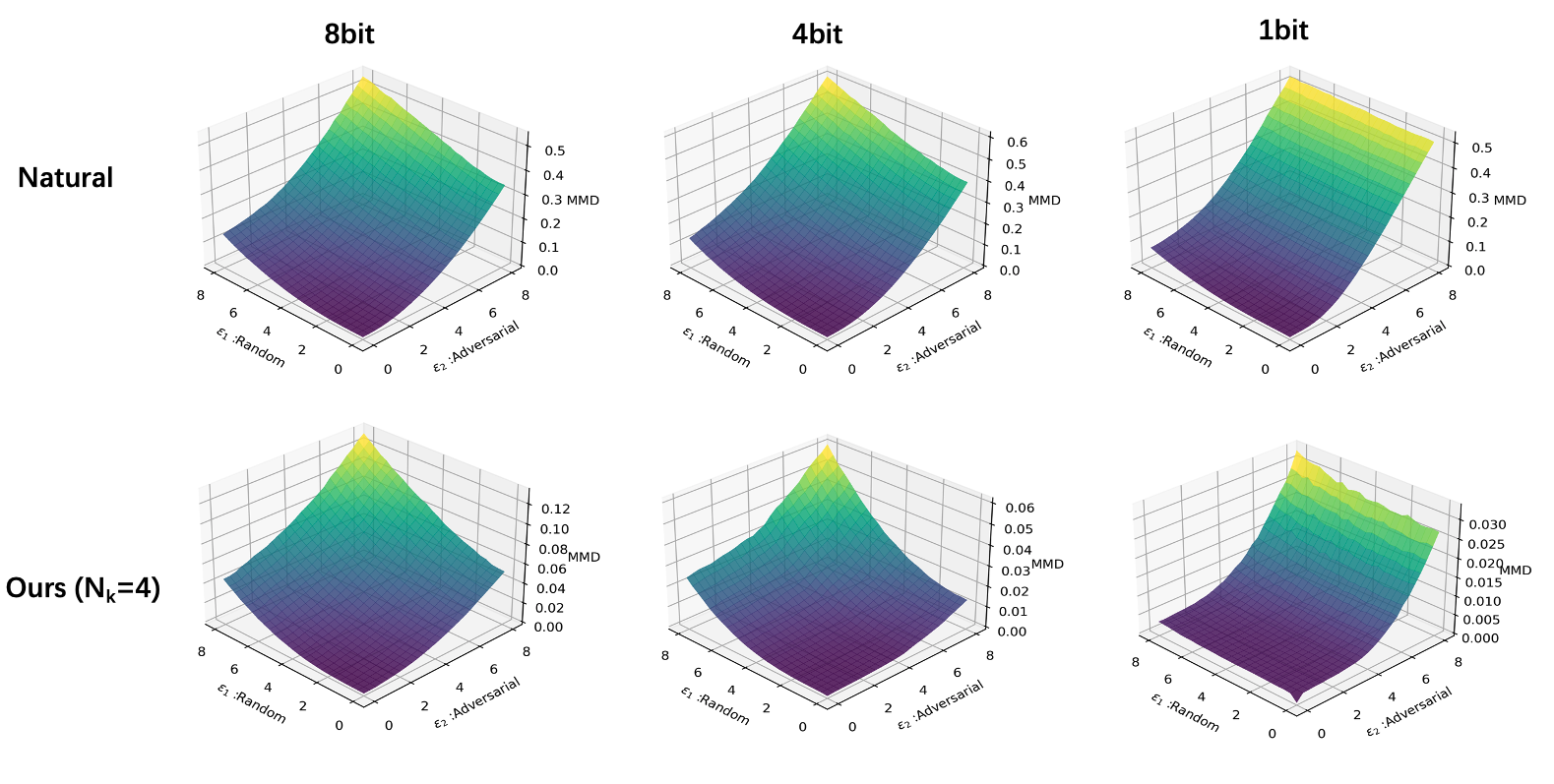}
\caption{Visualization of MMD loss surface in different bitwidths, which measures the MMD distance between the natural data (from domain $S_0$) and attacked data during inference (from domain $T$). The input image is perturbed on random direction $\epsilon_1$ and adversarial domain $\epsilon_2$.  } 
\label{fig:mmd}
\end{figure*}




\section{Implementation Details}
\label{ap:details}
In the experiments for the CIFAR-10 and MNIST datasets, we adopt ResNet-20 as the network architecture. The batch size is set to 512 with an initial learning rate of 0.1. The total epochs for natural training and ours are set as 300 epochs and 150 epochs, respectively. The learning rate decays to one-tenth of the original every 50 and 100 epochs for natural training and ours. The coefficient $\lambda$ is set to 3 by default, and is set to 0.003 for 1-bit experiment. For the attack methods GN, FGSM, PGD, BIM and TPGD involved in the experiments, we divide the clean images by 255 to scale them into the range $\left [ 0, 1 \right ]$ before sending into the neural network, and then the magnitude of perturbation $\epsilon$ is also divided by 255. The  perturbation alpha of small step size in PGD, BIM and TPGD are set as 4 with 20 iterations, which are used to generated PGD-20, BIM-20, TPGD-20 adversarial attacked data. We train experiments on CIFAR-10 and MNIST using one NVIDIA-3090 card. In the experiments for the ImageNet dataset, we adopt ResNet-18 as the network architecture. The total epochs for natural training and ours are set to 90 epochs and 45 epochs, respectively. The learning rate decays to one-tenth of the original every 30 and 15 epochs for natural training and ours. The schedule to generate attacked ImageNet data is consistent with that of CIFAR-10 and MNIST datasets. We train experiments on ImageNet with two NVIDIA-3090 cards.

\section{Visualization of the Loss Surfaces}

In Fig.~\ref{fig:cls_loss_surface}, we study the robustness against increasing disturbance in different bitwidth settings. We enlarge the perturbation gradually from 0 to 8, along the random direction $\epsilon_1$ and adversarial direction  $\epsilon_2$. Compared with natural training, our training reduces the loss scale  against the perturbation by a large margin, either in the random or adversarial direction.  In addition, the loss landscape is more rugged in the 1-bit setting, in which the model has only +1 or -1 representation after each binarized layer. For both weight and activation, the small perturbation may cause either error absorption or even the change of sign through each binarized layer, which leads to an uneven loss landscape. It illustrates why robustness improvement on the 1-bit network is the most challenging task, compared with other precision settings.

Fig.~\ref{fig:mmd} visualizes the MMD loss surface after training. The MMD distance shows the dissimilarity of the high-level feature distributions between the natural data and attacked data during inference. Large distance means dissimilar distribution. Compared with the loss surface with natural training (top line), the proposed approach enables  MMD distance reduction effectively, which explains why the proposed approach classifies attacked data samples correctly, even having comparable performance with the model on clean data in some cases.

\section{Analyses of the Number of Source Domains}

\begin{figure*}[hb]
\centering
\subfigure[$N_k=4$, Label:mud turtle]{
\begin{minipage}[t]{0.5\linewidth}
\centering
\includegraphics[width=2.8in, height=3.5in]{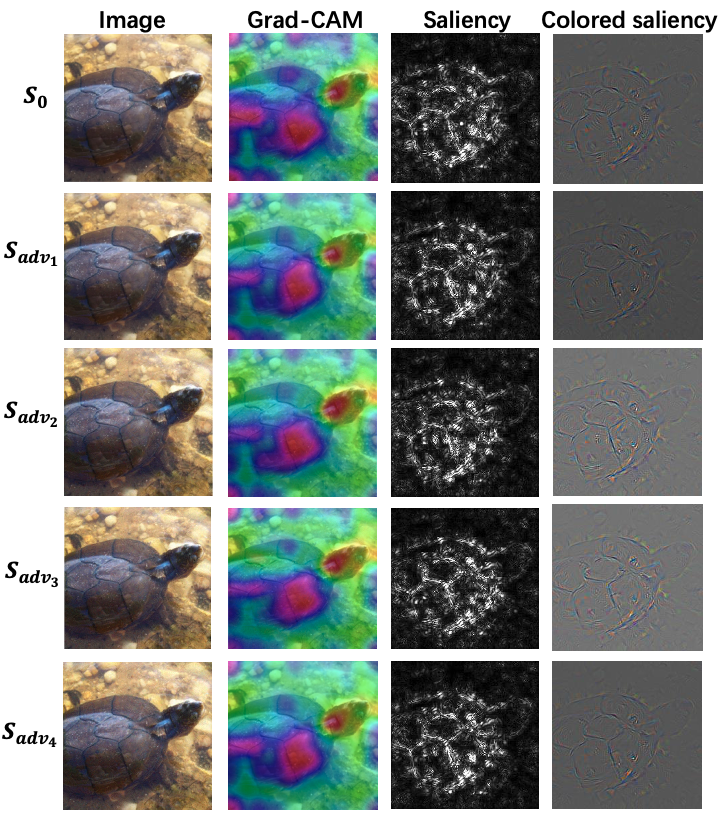}
\end{minipage}%
}%
\subfigure[$N_k=4$, Label:goose]{
\begin{minipage}[t]{0.5\linewidth}
\centering
\includegraphics[width=2.8in, height=3.5in]{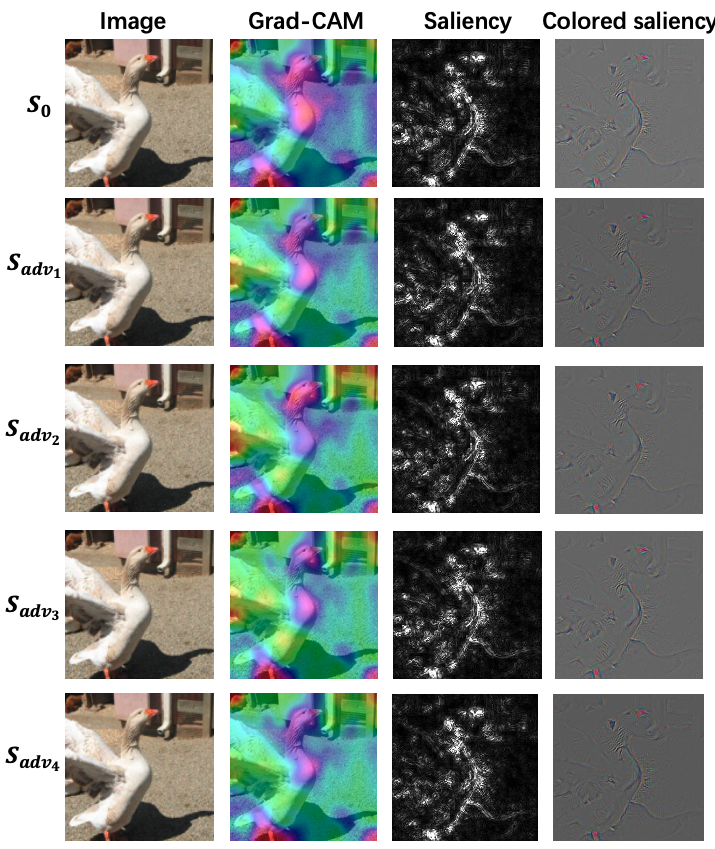}
\end{minipage}
}%

\subfigure[$N_k=2$, Label:water ouzel]{
\begin{minipage}[t]{0.5\linewidth}
\centering
\includegraphics[width=2.8in, height=2.1in]{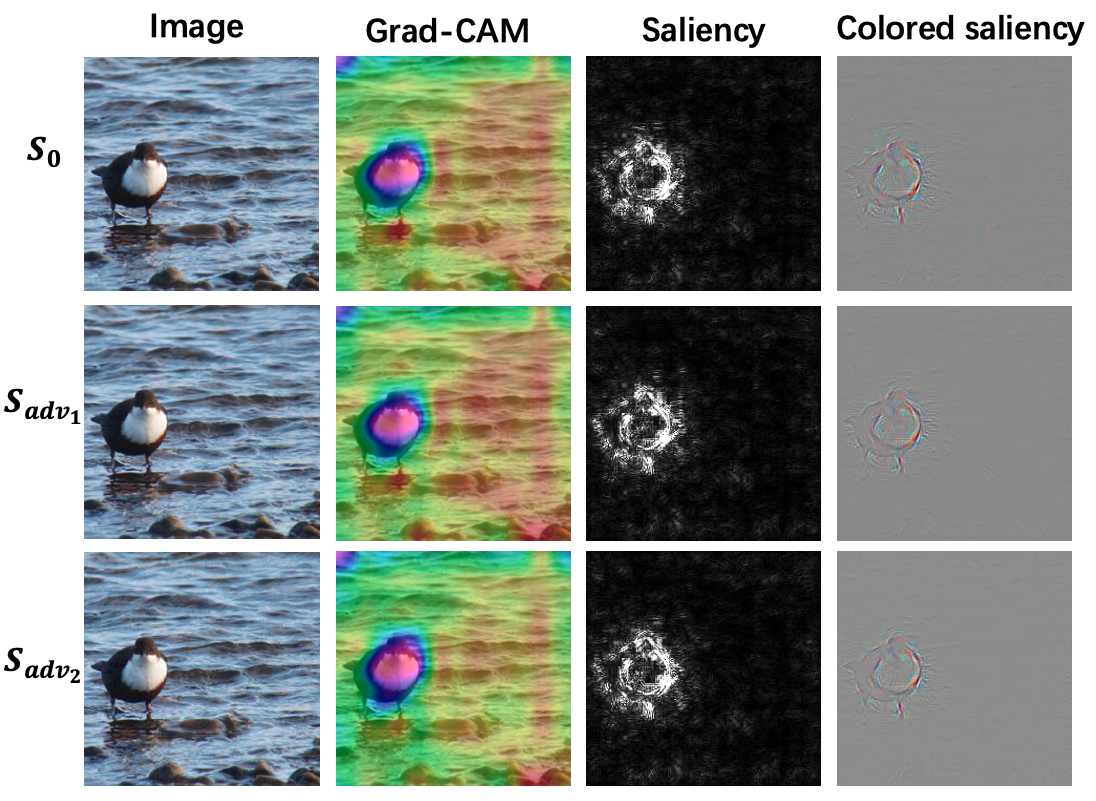}
\end{minipage}%
}%
\subfigure[$N_k=2$, Label:ostrich]{
\begin{minipage}[t]{0.5\linewidth}
\centering
\includegraphics[width=2.8in, height=2.1in]{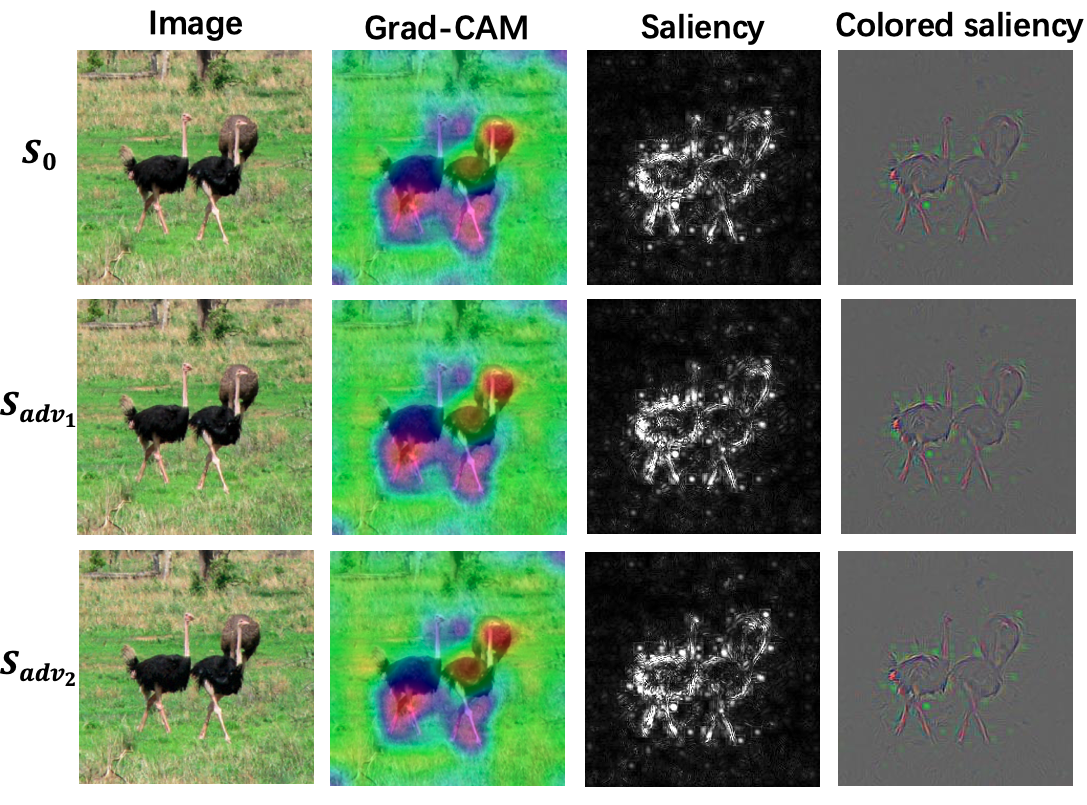}
\end{minipage}
}%
\caption{Visualizations to investigate what the model learns from gradient in different number of source domains. The gradient information is visualized by Grad-CAM and Back-propagation Saliency map, which reflect the gradient of the class score \textit{w.r.t} the input data. Fewer high-confidence areas without the region of interest (RoI) are given in adversarial domains than that in the natural domain $S_0$. It hints that training with adversarial samples makes the model attend more to the RoI, thereby improving robustness.}
\label{fig:more_vis}
\end{figure*}

More visualizations about the generated online adversarial data in Fig.~\ref{fig:more_vis}, where $N_k$ is the number of online adversarial domains. Both Grad-CAM and saliency map show that different gradient information is contained in data from different source domains. 

\end{document}